\documentclass[runningheads]{llncs}
\usepackage{caption}
\usepackage{hyperref}
\hypersetup{colorlinks=false, linkcolor=white, anchorcolor=white, citecolor=black}
\usepackage{graphicx}
\usepackage{subfig}
\usepackage{amsmath,amssymb,amsfonts}
\usepackage{booktabs}
\usepackage{epsfig} 
\begin{document}
\title{Semi-adaptive Synergetic Two-way Pseudoinverse Learning System}
%
%
\author{Binghong Liu\inst{1} \and
Ziqi Zhao\inst{1} \and
Shupan Li\inst{1,2} \and Ke Wang\inst{1,2}\thanks{Corresponding author: iekwang@zzu.edu.cn}}
%
%
\institute{School of Computer
and Artificial Intelligence, Zhengzhou University, Zhengzhou 450001, China. \and
National Supercomputing Center in Zhengzhou, Zhengzhou 450001, China.}
\maketitle              
\begin{abstract}
Deep learning has become a crucial technology for making breakthroughs in many fields. Nevertheless, it still faces two important challenges in theoretical and applied aspects. The first lies in the shortcomings of gradient descent based learning schemes which are time-consuming and difficult to determine the learning control hyperparameters. Next, the architectural design of the model is usually tricky. In this paper, we propose a semi-adaptive synergetic two-way pseudoinverse learning system, wherein each subsystem encompasses forward learning, backward learning, and feature concatenation modules. The whole system is trained using a non-gradient descent learning algorithm. It simplifies the hyperparameter tuning while improving the training efficiency. The architecture of the subsystems is designed using a data-driven approach that enables automated determination of the depth of the subsystems. We compare our method with the baselines of mainstream non-gradient descent based methods and the results demonstrate the effectiveness of our proposed method. The source code for this paper is available at \href{http://github.com/B-berrypie/Semi-adaptive-Synergetic-Two-way-Pseudoinverse-Learning-System}{http://github.com/B-berrypie/Semi-adaptive-Synergetic-Two-way-Pseudoinverse-Learning-System}. 

\end{abstract}
\section{Introduction}
Deep learning\cite{hao2016deep}, as a powerful representation learning technology, has achie-ved remarkable accomplishments in various domains, exerting profound influences on human society\cite{xu2020deep}. Gradient descent algorithm and its variants are a class of commonly used optimization algorithms employed to train deep neural networks by minimizing loss functions. The basic idea is to iteratively adjust parameters in the opposite direction of the gradient of the loss function, gradually approaching the minimum point of the function\cite{DBLP:journals/corr/Ruder16}. These algorithms are widely utilized in the field of deep learning due to their simplicity, ease of implementation and parallelizability\cite{DBLP:journals/jgo/GaoCWH23}. However, gradient descent algorithms face a range of challenges, including low training efficiency, the difficulty in setting hyperparameters, and the possibility of encountering issues like gradient vanishing and gradient explosion. Simultaneously, network architecture design and computational resource constraints represent two significant challenges in deep learning. 

To overcome the limitations of gradient descent algorithms, researchers have investigated many non-gradient descent methods. Extreme Learning Machine (ELM)\cite{1380068,DBLP:journals/mta/WangLWZ22} is a model based on a single-hidden layer feedforward neural network (SLFN). Unlike conventional gradient based algorithms such as backpropagation, ELM employs a unique training strategy: the input weights and biases are randomly assigned, and then the output weights are analytically determined using the Moore-Penrose generalized inverse. Hierarchical Extreme Learning Machine (HELM) \cite{DBLP:journals/tnn/TangDH16} is an extension of ELM, introducing a hierarchical structure to enhance the model's capability and performance. Broad learning system (BLS)\cite{7987745,9380770} draws inspiration from the random vector functional link neural network (RVFLNN)\cite{MALIK2023110377,DBLP:journals/ijon/PaoPS94}. BLS is configured as a flat network, where the initial inputs are embedded into feature nodes. The structure then undergoes broad expansion through enhancement nodes. 
The key idea of these representative non-gradient descent learning algorithms can also be traced back to the PseudoInverse Learning algorithm (PIL)\cite{ping1995An,DBLP:conf/inns/GuoZHF19} which was originally proposed for training SLFN.
In PIL, the output weights are calculated analytically by calculating an approximate optimal solution for the loss function using the Moore-Penrose generalized inverse. 
Several variants of the PIL algorithm\cite{DBLP:conf/isnn/WangG018} have been investigated that can use either the random weights or the pseudoinverse of the input data or its low-rank approximation as input weights. 

Motivated by the difficulty of designing network structures and the obstacles faced by gradient descent based learning algorithms, we propose a semi-adaptive synergetic two-way pseudoinverse learning system. Instead of pre-setting the model structures before training, the model grows dynamically according to the learning task during the training phase. 

Our contributions can be outlined as follows:
\begin{itemize}
\item We propose a synergetic learning system exhibiting superior performance contrasted with the baselines. Each elementary model of the proposed system includes two-way learning and feature fusion modules, enabling the acquisition of more comprehensive features.
\item The elementary model of the learning system is trained using a non-gradient approach, while the network architecture is dynamically determined, simplifying hyperparameter tuning.
\item The elementary training model within the synergetic learning system can be trained in parallel, facilitating the acceleration of the model training process. 
\end{itemize}
\section{Related Work}
\subsection{Pseudoinverse Learning based Autoencoder}
\begin{figure}[b]
      \centering
      \includegraphics[width=0.9\textwidth]{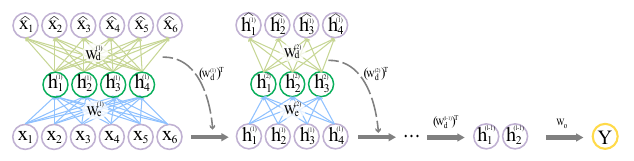}
      \caption{A schematic of the methodology to build stacked PILAE}
      \label{fig:stackPILAE}
\end{figure}
An autoencoder\cite{doi:10.1126/science.1127647}, a commonly used neural network employed in unsupervised learning, comprises an encoder and a decoder. The encoder transforms input data into a low-dimensional representation, while the decoder reconstructs this representation to an output closely resembling the original input. The loss function for an autoencoder can be defined as  
\begin{equation}
    \mathcal{L}(\mathbf{W}_{e},\mathbf{W}_{d}) = \frac{1}{2N}\sum_{i = 1}^{N}\left \| g(\mathbf{W}_{d}f(\mathbf{W}_{e}x_{i}-x_{i})) \right \|_{2}^{2},
\end{equation}
where $x_{i}$ denotes the i-th sample within the input data set, $N$ is the number of samples, while $f$ and $g$ represent the activation functions of the encoder and decoder, respectively. $\mathbf{W}_{e}$ denotes the encoder weight, and $\mathbf{W}_{d}$ denotes the decoder weight.

Given the challenges faced in training autoencoders using gradient descent methods, such as issues with hyperparameter tuning and inefficiency in training, the pseudoinverse learning based autoencoder (PILAE)\cite{DBLP:conf/smc/WangGXY17} was introduced. The setting of encoder weights $W_{e}$ varies across different versions of the PIL algorithm and its derivatives. Let $\textbf{H}$ represent the output of the encoder. The regularization term is used to prevent overfitting, and the new loss function is
\begin{equation}
    \mathcal{L}(\mathbf{W}_{d}) = \frac{1}{2}\left \| \mathbf{W}_{d}\mathbf{H}-\mathbf{X} \right \|_{2}^{2}+\frac{\lambda}{2}\left\| \mathbf{W}_{d} \right\|_{r}.
    \label{lossFReg}
\end{equation}
$\mathbf{X}\in \mathbb{R}^{d \times N}$ is the input data, d represents dimension. When $r$ is set to 2, the optimization problem associated with this loss function is also referred to as ridge regression. It can be readily inferred that 
\begin{equation}
    \mathbf{W}_{d} = \mathbf{X}\mathbf{H}^{T}(\mathbf{H}\mathbf{H}^{T}+\lambda \mathbf{I})^{-1}.
    \label{W_d}
\end{equation}
In autoencoder architectures, weight tying, where encoder weights are constrained to be equal to the transpose of decoder weights ($\mathbf{W}_{e} = \mathbf{W}_{d}^{T}$), is a common practice. This approach capitalizes on the symmetry inherent in autoencoders, enabling parameter sharing between encoder and decoder layers. 

Multiple PILAEs can be stacked to form a multilayer structure, facilitating the acquisition of increasingly abstract and sophisticated representations of the input data. Stacked PILAE typically employs a greedy layer-wise training approach, where each PILAE is trained independently. The output of the preceding PILAE's hidden layer serves as the input to the subsequent PILAE. 
\subsection{Synergetic Learning System}
\begin{figure}[t]
      \centering
      \includegraphics[width=0.6\textwidth]{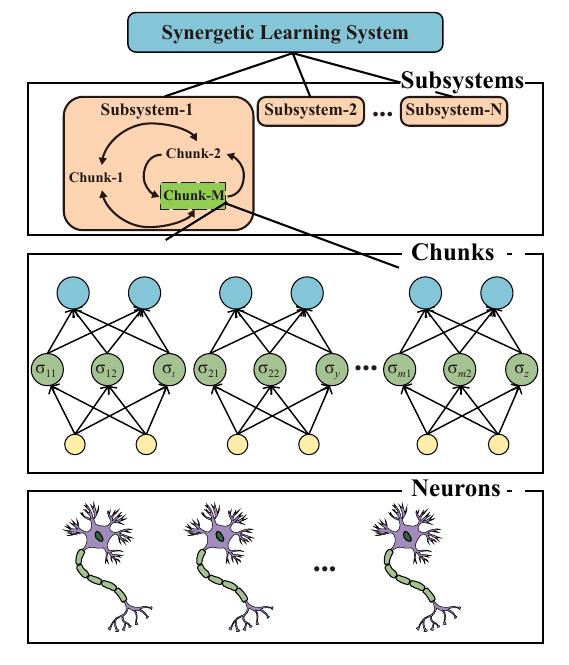}
      \caption{A diagram of the methodology to build SLS}
      \label{fig:sls}
   \end{figure}
   
A Synergetic Learning System (SLS)\cite{10091253,guo2020synergetic,9445740,XIA2022204} is a system that integrates at least two subsystems. Subsystems can be agents, models, or human-machine synergies. Synergetic learning between subsystems can be master and servant, cooperative or adversarial, as determined by the learning task. 

As shown in Fig.~\ref{fig:sls}, SLS can be built hierarchically, beginning with the assembly of individual neurons into functional units called chunks. These chunks are then integrated into larger subsystems, which combine to form the intricate architecture of neural networks. Finally, SLS emerges as a synergetic network comprising multiple interconnected subsystems, each contributing to the overall functionality of the system. 
\section{Methodology}
\begin{figure*}[t]
      \centering
      \includegraphics[width=1.0\textwidth]{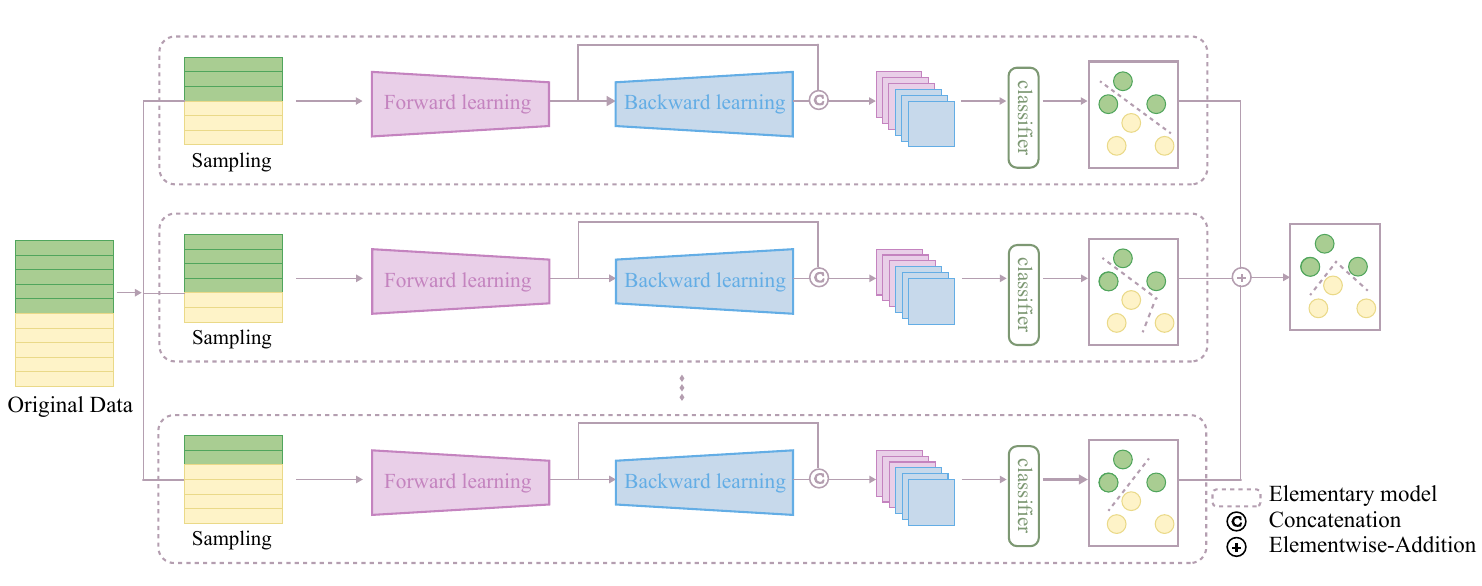}
      \caption{The structure of semi-adaptive synergetic two-way pseudoinverse learning system}
      \label{fig:mainDiagram}
\end{figure*}
Fig.~\ref{fig:mainDiagram} illustrates the architecture of our proposed method. The system utilizes an SLS framework consisting of several multi-level elementary models, with each elementary model being a hybrid neural network. This design allows the system to capture and process data at different levels, thereby enhancing the system's versatility and performance. 
\subsection{Elementary Model}
\begin{figure*}[thpb]
      \centering
      \includegraphics[width=1.0\textwidth]{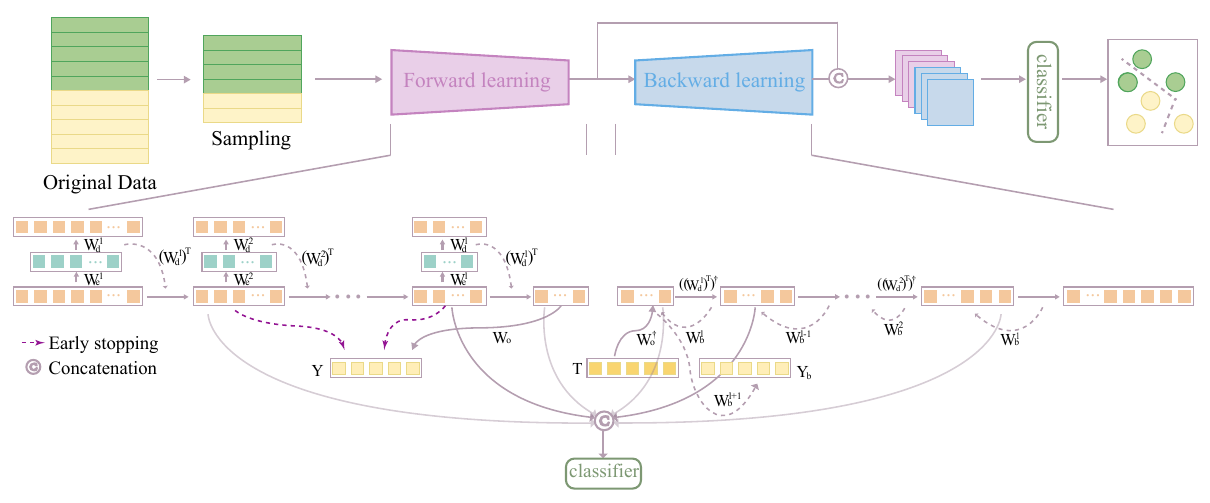}
      \caption{The structure of the elementary model}
      \label{fig:basemodel}
\end{figure*}

Each elementary model comprises three modules, forward learning, backward learning and concatenated feature fusion. Forward and backward learning work together to form two-way learning. The structure of the elementary model is illustrated in Fig.~\ref{fig:basemodel}. 
\subsubsection{Forward Learning} 
During the forward learning process, we utilize PILAE as the foundational block to construct a multi-layer network, namely stacked PILAE. As depicted in Eq.~(\ref{lossFReg}), $L_1$ regularization can be chosen when feature selection or achieving a sparse solution is desired, with the setting $r = 1$. This optimization problem is also termed as LASSO \cite{lasso}. To solve this optimization problem, various methods can be employed\cite{7102696}, including least angle regression (LAR)\cite{10.1214/08-SS035}, iterative shrinkage-thresholding algorithm (ISTA), and alternating direction method of multipliers (ADMM)\cite{Han2022ASO}. This study utilizes the fast iterative shrinkage-thresholding algorithm (FISTA)\cite{4959678,8578294}, an efficient variant of ISTA, to tackle the optimization problem. The main distinction between FISTA and ISTA lies in the selection of the starting point of the approximation function in the iteration step. Specifically, the basic iterative steps of FISTA are as follows
\begin{equation}
    \mathbf{W}_{d}^{k} = p_{L}(\mathbf{V}^{k}),
\end{equation}
\begin{equation}
    t_{k+1} = \frac{1+\sqrt{1+4t_{k}^{2}} }{2} ,
\end{equation}
\begin{equation}
    \mathbf{V}^{k+1} = \mathbf{W}_{d}^{k}+\left ( \frac{t_{k}-1}{t_{k+1}} \right ) \left ( \mathbf{W}_{d}^{k}-\mathbf{W}_{d}^{k-1} \right ),
\end{equation}
where $\mathbf{V}^{1} = \mathbf{W}_{d}^{0}$, $t_{1} = 1$. $L = L(q)$ is the Lipschitz constant of $\bigtriangledown q$. $p_{L}\left ( \cdot  \right ) $ denotes the proximal operator, defined as
\begin{equation}
    p_{L}\left( \mathbf{V}\right) = \underset{\mathbf{W}_d }{argmin}\left \{ \left \| \mathbf{W}_{d} \right \|_{1}+\frac{L}{2} \left \| \mathbf{W}_{d}-(\mathbf{V}-\frac{1}{L}\bigtriangledown q(\mathbf{V})) \right \|_{2}^{2} \right \},
\end{equation}
where $q(\mathbf{V}) = \left \| \mathbf{VH}-\mathbf{X} \right \|_{2}^{2} $, $\bigtriangledown q = 2E_{max}(\mathbf{H}\mathbf{H}^{T})$, $E_{max}(\cdot)$ computes the maximum eigenvalue of the given matrix. 
Through tied weights, the parameters of the encoder can be determined post-training. Removing the decoder, the output of the encoder is fed as input to the next PILAE, iteratively forming a stacked PILAE. The forward propagation function for forward learning can be expressed as 
\begin{equation}
    F(\mathbf{X}) = \sigma(\mathbf{W}_{e}^{l}\sigma(\mathbf{W}_{e}^{l-1}...\sigma(\mathbf{W}_{e}^{2}\sigma(\mathbf{W}_{e}^{1}\mathbf{X}))...)),
    \label{feedforwardF}
\end{equation} 
where $\sigma(\cdot)$ is the activation function. According to Eq.~(\ref{W_d}), it can be inferred that
\begin{equation}
    \mathbf{W}_{e}^{l} = (\mathbf{H}^{l-1}(\mathbf{H}^{l})^{T}(\mathbf{H}^{l}(\mathbf{H}^{l})^{T}+\lambda\mathbf{I})^{-1})^{T}. 
\end{equation}
\begin{equation}
    \mathbf{H}^{l}= \sigma(\mathbf{W}_{e}^{l}\sigma(\mathbf{W}_{e}^{l-1}...\sigma(\mathbf{W}_{e}^{2}\sigma(\mathbf{W}_{e}^{1}\mathbf{X}))...)). 
\end{equation} 
In particular, $\mathbf{H}^{0} = \mathbf{X}$. 

As shown in Fig.~\ref{fig:stackPILAE}, according to Eq.~(\ref{feedforwardF}) the final output of the task is represented as
\begin{equation}
    \mathbf{Y} = \mathbf{W}_{o}F(\mathbf{X}),
\end{equation}
where $\mathbf{Y} $ represents the output, the weight matrix $\mathbf{W}_{o}$ connects the last hidden layer to the output layer. The objective is to minimize the discrepancy between the output $\mathbf{Y}$ and the real label matrix $\mathbf{T}$. 
\begin{equation}
    minimize\left \| \mathbf{Y}-\mathbf{T} \right \| _{2}^{2} = minimize\left \| \mathbf{W}_{o}F(\mathbf{X} )-\mathbf{T}\right\|.
    \label{minimizeF}
\end{equation}
According to Eq.~(\ref{lossFReg}) and Eq.~(\ref{W_d}), $\mathbf{W}_{o}$ can be computed as
\begin{equation}
    \mathbf{W}_{o} = \mathbf{T}F(\mathbf{X} )^{T}(F(\mathbf{X} )F(\mathbf{X} )^{T}+\lambda\mathbf{I})^{-1}.
\end{equation}

In deep learning, determining the optimal depth of a model is a challenging task, so building the model incrementally to meet the task requirements is a sensible strategy. Model construction begins with a simpler and shallower architecture, gradually increasing its complexity as needed. 
\subsubsection{Backward Learning}
Forward learning, which primarily relies on unsupervised data reconstruction tasks for feature extraction, inevitably leads to features that are more suited for reconstruction tasks, thus potentially hindering performance on downstream learning tasks. Additionally, forward learning fails to fully leverage the information contained in target labels which is evidently beneficial for extracting features associated with specific learning tasks. This prompts us to propose backward learning. 
Once the training of the forward learning network is completed, the structure of the network is established. Backward learning employs the same architecture, learning in reverse from the last hidden layer back to the input layer. 

According to Eq.~(\ref{minimizeF}), the $\mathbf{H}_{b}^{l}$ that yields the minimum error can be determined as follows: 
\begin{equation}
    \mathbf{H}_{b}^{l} = \mathbf{W}_{o}^{\dagger}\mathbf{T}.
    \label{backHl}
\end{equation}
$\dagger$ denotes pseudoinverse (Moore-Penrose inverse).
Similarly,
\begin{equation}
    \mathbf{H}_{b}^{l-1} = (\mathbf{W}_{e}^{l})^{\dagger}\sigma^{-1}(\mathbf{H}_{b}^{l}).
    \label{backfH}
\end{equation}
$\sigma^{-1}(\cdot)$ represents the inverse function of $\sigma(\cdot)$. 

Repeating Eq.~(\ref{backfH}) yields:
\begin{equation}
    \mathbf{H}_{b}^{1} =(\mathbf{W}_{e}^{2})^{\dagger}\sigma^{-1}\left(... (\mathbf{W}_{e}^{l-1})^{\dagger}\sigma^{-1}\left((\mathbf{W}_{e}^{l})^{\dagger}\sigma^{-1}(\mathbf{H}_{b}^{l}) \right)...\right).
    \label{backH1}
\end{equation}
Substituting Eq.~(\ref{backHl}) into Eq.~(\ref{backH1}),
\begin{equation}
    \mathbf{H}_{b}^{1} =(\mathbf{W}_{e}^{2})^{\dagger}\sigma^{-1}\left(... (\mathbf{W}_{e}^{l})^{\dagger}\sigma^{-1}(\mathbf{W}_{o}^{\dagger}\mathbf{T})...\right).
\end{equation}
At this stage, the label information propagates backward through the network. The reconstructed hidden layer output contains features related to the label. The weights undergo corresponding updates. Like Eq.~(\ref{minimizeF}), the weights are updated by
\begin{equation}
    minimize\left\| \mathbf{W}_{b}^{1}\mathbf{X}-\mathbf{H}_{b}^{1} \right\|_{2}^{2}, 
\end{equation}
then, 
\begin{equation}
    \mathbf{W}_{b}^{1} = \mathbf{H}_{b}^{1}\mathbf{X}^{\dagger}, 
\end{equation}
similarly,
\begin{equation}
    \mathbf{W}_{b}^{2} = \mathbf{H}_{b}^{2}\left(\sigma(\mathbf{W}_{b}^{1}\mathbf{X})\right)^{\dagger}. 
\end{equation}
\begin{equation}
    \mathbf{W}_{b}^{l} = \mathbf{H}_{b}^{l}\left(\sigma(\mathbf{W}_{b}^{l-1}\sigma(...\sigma(\mathbf{W}_{b}^{1}\mathbf{X})...))\right)^{\dagger}.
\end{equation}
Let $\mathbf{W}_{b}^{l+1}$ denote the output weight.
\begin{equation}
    \mathbf{W}_{b}^{l+1} = \mathbf{T}\left(\sigma(\mathbf{W}_{b}^{l}\sigma(...\sigma(\mathbf{W}_{b}^{1}\mathbf{X})...))\right)^{\dagger}.
\end{equation}
After determining the weights through backward learning, the prediction of the backward learning network can be expressed as
\begin{equation}
    \mathbf{Y}_{b} = \mathbf{W}_{b}^{l+1} \sigma(\mathbf{W}_{b}^{l}\sigma(\mathbf{W}_{b}^{l-1}...\sigma(\mathbf{W}_{b}^{2}\sigma(\mathbf{W}_{b}^{1}\mathbf{X}))...)).
\end{equation}

\subsubsection{Feature Fusion}
Integrating features from diverse sources enables the acquisition of richer information, facilitating a more profound understanding of the data and the extraction of underlying patterns, thus enhancing the robustness of the model. 

Forward and backward learning will be used as a representation learning network instead of directly performing the final learning task. We concatenate features of varying abstraction levels obtained from forward and backward learning for downstream tasks. For networks with $l$ hidden layers, there are multiple fusion methods. When selecting one feature from the forward learning path and one from the backward learning path for concatenation, there are $l \times l$ possible combinations. Alternatively, fusion can involve selecting multiple features from both the forward and backward paths. 
\subsection{Synergetic System}
In our work, each elementary model functions as a subsystem within the larger synergetic system. These subsystems operate cooperatively to accomplish the designated task. 
Each subsystem is equipped with a two-way training model and a feature fusion module, enhancing the system's adaptability and performance. The synergistic interaction and complementarity among these subsystems contribute significantly to the overall system's remarkable classification accuracy. Furthermore, the modular design of the subsystems facilitates their deployment in a distributed manner, resulting in a substantial reduction in computational time overhead. 
\subsection{Training Strategy}
\subsubsection{Parallelizability}
In our method, each elementary model is capable of independently completing its task, allowing elementary models to be trained simultaneously, thus enabling parallelization. During training, each elementary model randomly samples a subset of the training set according to a predefined sampling ratio. This approach fosters diversity and heterogeneity among the elementary models, which in turn enhances the overall stability and generalization performance of the synergetic learning system. 
\subsubsection{Early Stopping}
Early stopping is a technique used to prevent overfitting during model training. Its basic principle involves monitoring the performance metrics of the model on a validation set, the training is halted with a high probability to prevent the model from continuing to overfit the training data. In our work, it is used as a structural control scheme.
\section{Experiments}
\subsection{Data Sets}
In our study, we rigorously evaluate the performance of proposed methodologies by employing a wide array of benchmark data sets that are widely recognized within the research community. Specifically, we include classical and contemporary data sets such as MNIST, Fashion-MNIST (F-MNIST), and the NORB data set \cite{LeCunNORB}. Additionally, we utilize a varied selection of data sets from the UCI Machine Learning Repository (\href{http://archive.ics.uci.edu/ml}{http://archive.ics.uci.edu/ml}) and the OpenML platform (\href{http://openml.org}{http://openml.org}), which have been commonly used in related works. These data sets include but are not limited to, Abalone, Advertisement, Gisette, the Human Activity Recognition (HAR) data set, the Kin8nm data set, Madelon, Mfeat, Isolet, Occupancy, Yeast, Semeion and Spambase.

\subsection{Experimental Design}
The performance of our method was compared to five baselines that included HELM \cite{DBLP:journals/tnn/TangDH16}, PILAE \cite{guo2021pilae}, ELM-AE \cite{6937189}, PILLS \cite{DBLP:journals/cogcom/WangG21}, and BLS \cite{7987745}. These methods use the non-gradient learning strategy. The primary distinction between them lies in the mapping linked to the input weights. They use least squares or ridge regression for output weight learning. 
For the MNIST, F-MNIST, and NORB data sets, the structural settings of the baselines take the optimal structure mentioned in the corresponding papers. 

In addition, we compared the training efficiency of our method with three typical gradient descent based methods (LeNet-5 \cite{726791}, ResNet50 \cite{DBLP:conf/cvpr/HeZRS16} and VGG16 \cite{DBLP:journals/corr/SimonyanZ14a}) on MNIST, F-MNIST and NORB. 

LeNet-5 comprises two convolutional layers with 6 and 16 $5 \times 5$ filters respectively, followed by two max-pooling layers. It has three fully connected layers with 120, 84, and 10 (output) nodes. LeNet-5 was trained using the Adam optimizer, with an initial learning rate of 0.001, batch size of 64, for 30 epochs. ResNet50 is a 50-layer deep residual network constructed from bottleneck residual blocks. It employs $1 \times 1$ and $3 \times 3$ convolutional filters, with skip connections to facilitate training of such a deep architecture. For ResNet50, we employed the Adam optimizer with an initial learning rate of 0.1 and a batch size of 64. The network was trained for 20 epochs. VGG16 utilizes very small $3 \times 3$ convolutional filters, with a substantially increased depth of 16 weight layers. It comprises five max-pooling layers and two fully-connected layers with 4096 nodes each. VGG16 was originally trained using stochastic gradient descent (SGD) with an initial learning rate of 0.001, batch size of 64, for 5 epochs.

All experiments were conducted on a PC equipped with an Intel(R) Core(TM) i5-14600K 3.50 GHz processor and 48.0 GB of DDR5 RAM. 
\subsection{Result and Analysis}

\begin{table}[h]
  \centering
  \caption{Comparison of Classification Accuracy between Our Method and Baselines }
    \begin{tabular}{lcccccc}
    \toprule
    Data sets & HELM  & PILAE & ELM-AE & PILLS & BLS   & Ours \\
    \midrule
    Abalone& 57.62 & 59.36 & 56.52 & 59.84 & 56.24 & \textbf{62.12} \\
    Advertisement & 96.94 & 95.42 & 93.77 & 96.73 & 91.76 & \textbf{98.34} \\
    Gisette & 97.60 & 97.22 & \textbf{97.88} & 97.33 & \textbf{97.88} & 97.60 \\
    Gina\_agnostic & 87.17 & 84.70 & 88.60 & 88.22 & 84.62 & \textbf{90.11} \\
    Har   & 83.88 & 95.95 & 96.68 & \textbf{97.35} & 82.76 & 96.86 \\
    Kin8nm & 79.77 & 80.44 & 77.53 & 85.82 & 81.65 & \textbf{88.27} \\
    Madelon & 72.36 & 56.70 & 71.90 & 58.25 & 64.13 & \textbf{87.64} \\
    Mfeat & 96.00 & 98.75 & 95.13 & 98.49 & 99.33 & \textbf{99.50} \\
    Isolet & 96.22 & 92.69 & 96.49 & 94.80 & 96.67 & \textbf{98.30} \\
    Occupancy & 96.56 & 97.47 & 98.54 & 98.54 & 97.45 & \textbf{98.98} \\
    Prior & 94.81 & 89.32 & 95.67 & 91.51 & 93.37 & \textbf{97.51} \\
    Yeast & 61.62 & 55.66 & 58.83 & 56.56 & 61.71 & \textbf{63.70} \\
    Semeion & 95.26 & 94.94 & 95.14 & 94.51 & 94.93 & \textbf{97.33} \\
    Segment & 92.32 & 95.39 & 93.01 & 89.28 & 92.86 & \textbf{95.48} \\
    Spambase & 90.19 & 91.09 & 85.83 & 91.13 & 86.14 & \textbf{93.27} \\
    Sylva  & 98.82 & 98.61 & 98.92 & 98.66 & 99.04 & \textbf{99.30} \\
    MNIST & 98.78 & 96.93 & 98.78 & 97.48 & 98.60 & \textbf{98.97} \\
    NORB  & 88.70 & 88.11 & 89.14 & 90.46 & 86.79 & \textbf{91.50} \\
    F-MNIST & 88.53 & 86.32 & 88.60 & 87.29 & \textbf{89.60} & 89.58 \\
    \bottomrule
    \end{tabular}%
  \label{tab:accuracy}%
\end{table}%

Our proposed method and baselines are compared on 19 public data sets in terms of accuracy on the test set, and the comparison results are shown in Table~\ref{tab:accuracy}. Overall, our performance outperformed the baselines in 16 out of 19 data sets. Specifically, our results show significant improvement over the baselines in Abalone, Advertisement, Gina\_agnostic, Kin8nm, Madelon, Isolet, Prior, Yeast, Semeion, Spambase, and NORB data sets. For Gisette, our method achieved the third best result, trailing only 0.28 behind the top-performing BLS and ELM-AE methods. For Har, our method achieves the second best result, trailing PILLS by 0.49. For F-MNIST, our method trailed the top-performing method by only 0.02. For MNIST, we are 0.19 ahead of the two best results of baselines. 

For the NORB data set, we set the forward learning structure to 1500-1000-600-500, which was the maximum number of hidden layers, and the actual number was decided using the early stopping strategy. The backward learning structure was the same as the structure determined by forward learning. The number of subsystems was set to 10, and the sampling ratio of each subsystem was 0.8. The number of classifier neurons for the fused feature was set to 5000. For F-MNIST, three subsystems were used, the sampling ratio of each subsystem was set to 0.8, the forward learning structure was 1500-1000-600-500, an early stopping strategy was used, and the number of classifier neurons for the fused features was set to 10000. For MNIST, two subsystems were used, the sampling ratio of each subsystem was set to 0.8, the forward learning structure was 2000-1500-500, using an early stopping strategy, and the number of classifier neurons for fused features was set to 10000.

\begin{figure}[t]
      \centering
      \includegraphics[width=0.5\textwidth]{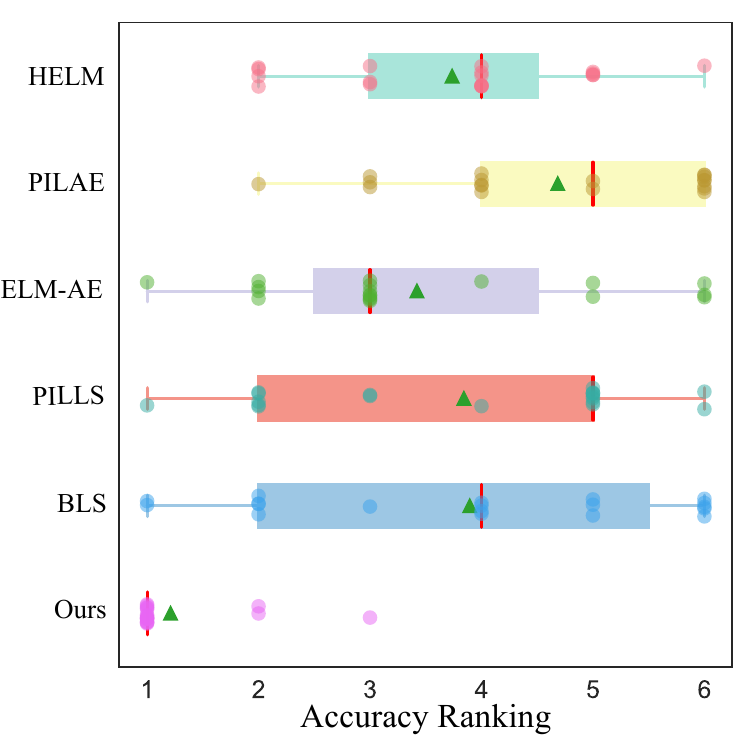}
      \caption{The accuracy ranking box plots of our method and baselines}
      \label{fig:boxplot}
\end{figure}
Fig.~\ref{fig:boxplot} shows the comparison of our method with the baselines in terms of ranking of accuracy in the test set. The red vertical line indicates the median and the green triangle indicates the mean. Obviously, the method we propose is leading the way. 


\begin{figure}[t]
    \centering
    \subfloat[Comparison on \\MNIST]{\includegraphics[width=0.33\textwidth]{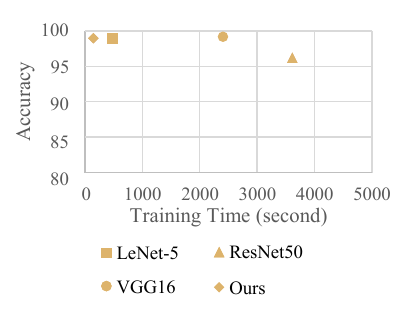}}
 \hfill 	
  \subfloat[Comparison on \\\centering F-MNIST]{\includegraphics[width=0.33\textwidth]{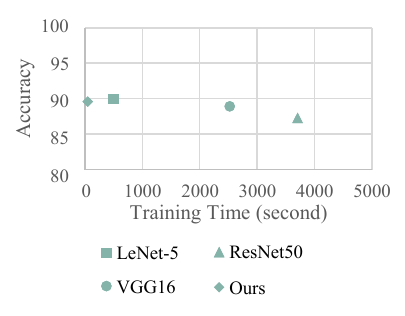}}
 \hfill	
  \subfloat[Comparison on \\NORB]{\includegraphics[width=0.33\textwidth]{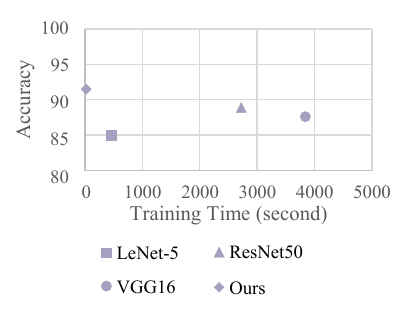}}
  \caption{Comparison of training efficiency between our method and gradient based methods}
      \label{fig:efficiency}
\end{figure}
A comparison of training efficiency is shown in Fig.~\ref{fig:efficiency}, our method achieves higher efficiency in the case that we attain approximate accuracy with the baselines. The accuracy of the baselines in the experiment may not have reached its optimal result because of the insufficient epoch setting, but the training time consumed has been significantly more than our method. 
\section{Conclusion}
In this work, we propose a semi-adaptive synergetic two-way pseudoinverse learning system, comprised of subsystems that interact synergistically. Each subsystem consists of three modules: forward learning, backward learning, and feature fusion. The contributions of this work are threefold. Firstly, experimental results demonstrate that our method exhibits superior performance compared to the representative competing baselines. Forward and backward learning constitute two-way learning that enables learning of richer features. Secondly, our method utilizes a non-gradient descent algorithm for training and employs a data-driven semi-adaptive strategy to determine the model structure, thereby reducing the workload for hyperparameter tuning. Additionally, the elementary models within the synergetic system can be trained in parallel, further enhancing the training efficiency. In fact, the forward and backward learning can also be viewed as subsystems in a synergetic system. Such nested synergetic learning systems will be further investigated in future work. 
\section*{Acknowledgement}
This study is supported by China Postdoctoral Science Foundation (2020M682348), National Key Research and Development Program of China (2018AAA0100203), and Natural Science Foundation of Henan Province, China (232300421235).

\bibliographystyle{splncs04}
\bibliography{mybibliography}

\begin{thebibliography}{10}
\providecommand{\url}[1]{\texttt{#1}}
\providecommand{\urlprefix}{URL }
\providecommand{\doi}[1]{https://doi.org/#1}

\bibitem{4959678}
Beck, A., Teboulle, M.: A fast iterative shrinkage-thresholding algorithm with
  application to wavelet-based image deblurring. In: 2009 IEEE International
  Conference on Acoustics, Speech and Signal Processing. pp. 693--696 (2009).
  \doi{10.1109/ICASSP.2009.4959678}

\bibitem{7987745}
Chen, C.L.P., Liu, Z.: Broad learning system: An effective and efficient
  incremental learning system without the need for deep architecture. IEEE
  Transactions on Neural Networks and Learning Systems  \textbf{29}(1),  10--24
  (2018). \doi{10.1109/TNNLS.2017.2716952}

\bibitem{DBLP:journals/jgo/GaoCWH23}
Gao, X., Cai, X., Wang, X., Han, D.: An alternating structure-adapted bregman
  proximal gradient descent algorithm for constrained nonconvex nonsmooth
  optimization problems and its inertial variant. J. Glob. Optim.
  \textbf{87}(1),  277--300 (2023). \doi{10.1007/S10898-023-01300-0},
  \url{https://doi.org/10.1007/s10898-023-01300-0}

\bibitem{9380770}
Gong, X., Zhang, T., Chen, C.L.P., Liu, Z.: Research review for broad learning
  system: Algorithms, theory, and applications. IEEE Transactions on
  Cybernetics  \textbf{52}(9),  8922--8950 (2022).
  \doi{10.1109/TCYB.2021.3061094}

\bibitem{guo2021pilae}
Guo, P., Wang, K., Zhou, X.L.: Pilae: A non-gradient descent learning scheme
  for deep feedforward neural networks (2021)

\bibitem{ping1995An}
Guo, P., Chen, C.L.P., Sun, Y.: An exact supervised learning for a three-layer
  supervised neural network. In: the International Conference on Neural
  Information Processing {ICONIP} 1995, Beijing, China, October 30-November 3,
  1995. pp. 1041--1044. Publishing House Of Electronics Industry (1995)

\bibitem{10091253}
Guo, P., Hou, J., Zhao, B.: Methodologies of building synergetic learning
  systems. In: 2022 18th International Conference on Computational Intelligence
  and Security (CIS). pp. 200--204 (2022). \doi{10.1109/CIS58238.2022.00049}

\bibitem{guo2020synergetic}
Guo, P., Yin, Q.: Synergetic learning systems: Concept, architecture, and
  algorithms (2020)

\bibitem{DBLP:conf/inns/GuoZHF19}
Guo, P., Zhao, D., Han, M., Feng, S.: Pseudoinverse learners: New trend and
  applications to big data. In: Oneto, L., Navarin, N., Sperduti, A., Anguita,
  D. (eds.) Recent Advances in Big Data and Deep Learning, Proceedings of the
  {INNS} Big Data and Deep Learning Conference {INNSBDDL} 2019, held at Sestri
  Levante, Genova, Italy 16-18 April 2019. pp. 158--168. Springer (2019).
  \doi{10.1007/978-3-030-16841-4\_17},
  \url{https://doi.org/10.1007/978-3-030-16841-4\_17}

\bibitem{Han2022ASO}
Han, D.R.: A survey on some recent developments of alternating direction method
  of multipliers. Journal of the Operations Research Society of China pp. 1--52
  (2022), \url{https://api.semanticscholar.org/CorpusID:245617402}

\bibitem{hao2016deep}
Hao, X., Zhang, G., Ma, S.: Deep learning. International Journal of Semantic
  Computing  \textbf{10}(03),  417--439 (2016)

\bibitem{DBLP:conf/cvpr/HeZRS16}
He, K., Zhang, X., Ren, S., Sun, J.: Deep residual learning for image
  recognition. In: 2016 {IEEE} Conference on Computer Vision and Pattern
  Recognition, {CVPR} 2016, Las Vegas, NV, USA, June 27-30, 2016. pp. 770--778.
  {IEEE} Computer Society (2016). \doi{10.1109/CVPR.2016.90},
  \url{https://doi.org/10.1109/CVPR.2016.90}

\bibitem{10.1214/08-SS035}
Hesterberg, T., Choi, N.H., Meier, L., Fraley, C.: {Least angle and $l_1$
  penalized regression: A review}. Statistics Surveys  \textbf{2}(none),  61 --
  93 (2008). \doi{10.1214/08-SS035}, \url{https://doi.org/10.1214/08-SS035}

\bibitem{doi:10.1126/science.1127647}
Hinton, G.E., Salakhutdinov, R.R.: Reducing the dimensionality of data with
  neural networks. Science  \textbf{313}(5786),  504--507 (2006).
  \doi{10.1126/science.1127647},
  \url{https://www.science.org/doi/abs/10.1126/science.1127647}

\bibitem{1380068}
Huang, G.B., Zhu, Q.Y., Siew, C.K.: Extreme learning machine: a new learning
  scheme of feedforward neural networks. In: 2004 IEEE International Joint
  Conference on Neural Networks (IEEE Cat. No.04CH37541). vol.~2, pp. 985--990
  vol.2 (2004). \doi{10.1109/IJCNN.2004.1380068}

\bibitem{726791}
Lecun, Y., Bottou, L., Bengio, Y., Haffner, P.: Gradient-based learning applied
  to document recognition. Proceedings of the IEEE  \textbf{86}(11),
  2278--2324 (1998). \doi{10.1109/5.726791}

\bibitem{LeCunNORB}
LeCun, Y., Huang, F.J., Bottou, L.: Learning methods for generic object
  recognition with invariance to pose and lighting. In: Proceedings of the IEEE
  Conference on Computer Vision and Pattern Recognition (2004)

\bibitem{MALIK2023110377}
Malik, A., Gao, R., Ganaie, M., Tanveer, M., Suganthan, P.N.: Random vector
  functional link network: Recent developments, applications, and future
  directions. Applied Soft Computing  \textbf{143},  110377 (2023).
  \doi{https://doi.org/10.1016/j.asoc.2023.110377},
  \url{https://www.sciencedirect.com/science/article/pii/S1568494623003952}

\bibitem{DBLP:journals/ijon/PaoPS94}
Pao, Y., Park, G.H., Sobajic, D.J.: Learning and generalization characteristics
  of the random vector functional-link net. Neurocomputing  \textbf{6}(2),
  163--180 (1994). \doi{10.1016/0925-2312(94)90053-1},
  \url{https://doi.org/10.1016/0925-2312(94)90053-1}

\bibitem{DBLP:journals/corr/Ruder16}
Ruder, S.: An overview of gradient descent optimization algorithms. CoRR
  \textbf{abs/1609.04747} (2016), \url{http://arxiv.org/abs/1609.04747}

\bibitem{DBLP:journals/corr/SimonyanZ14a}
Simonyan, K., Zisserman, A.: Very deep convolutional networks for large-scale
  image recognition. In: Bengio, Y., LeCun, Y. (eds.) 3rd International
  Conference on Learning Representations, {ICLR} 2015, San Diego, CA, USA, May
  7-9, 2015, Conference Track Proceedings (2015),
  \url{http://arxiv.org/abs/1409.1556}

\bibitem{DBLP:journals/tnn/TangDH16}
Tang, J., Deng, C., Huang, G.: Extreme learning machine for multilayer
  perceptron. {IEEE} Trans. Neural Networks Learn. Syst.  \textbf{27}(4),
  809--821 (2016). \doi{10.1109/TNNLS.2015.2424995},
  \url{https://doi.org/10.1109/TNNLS.2015.2424995}

\bibitem{lasso}
Tibshirani, R.: Regression shrinkage and selection via the lasso. Journal of
  the Royal Statistical Society: Series B (Methodological)  \textbf{58}(1),
  267--288 (1996). \doi{https://doi.org/10.1111/j.2517-6161.1996.tb02080.x},
  \url{https://rss.onlinelibrary.wiley.com/doi/abs/10.1111/j.2517-6161.1996.tb02080.x}

\bibitem{DBLP:journals/mta/WangLWZ22}
Wang, J., Lu, S., Wang, S., Zhang, Y.: A review on extreme learning machine.
  Multim. Tools Appl.  \textbf{81}(29),  41611--41660 (2022).
  \doi{10.1007/S11042-021-11007-7},
  \url{https://doi.org/10.1007/s11042-021-11007-7}

\bibitem{DBLP:conf/isnn/WangG018}
Wang, J., Guo, P., Xin, X.: Review of pseudoinverse learning algorithm for
  multilayer neural networks and applications. In: Huang, T., Lv, J., Sun, C.,
  Tuzikov, A.V. (eds.) Advances in Neural Networks - {ISNN} 2018 - 15th
  International Symposium on Neural Networks, {ISNN} 2018, Minsk, Belarus, June
  25-28, 2018, Proceedings. Lecture Notes in Computer Science, vol. 10878, pp.
  99--106. Springer (2018). \doi{10.1007/978-3-319-92537-0\_12},
  \url{https://doi.org/10.1007/978-3-319-92537-0\_12}

\bibitem{DBLP:journals/cogcom/WangG21}
Wang, K., Guo, P.: A robust automated machine learning system with
  pseudoinverse learning. Cogn. Comput.  \textbf{13}(3),  724--735 (2021).
  \doi{10.1007/S12559-021-09853-6},
  \url{https://doi.org/10.1007/s12559-021-09853-6}

\bibitem{DBLP:conf/smc/WangGXY17}
Wang, K., Guo, P., Xin, X., Ye, Z.: Autoencoder, low rank approximation and
  pseudoinverse learning algorithm. In: 2017 {IEEE} International Conference on
  Systems, Man, and Cybernetics, {SMC} 2017, Banff, AB, Canada, October 5-8,
  2017. pp. 948--953. {IEEE} (2017). \doi{10.1109/SMC.2017.8122732},
  \url{https://doi.org/10.1109/SMC.2017.8122732}

\bibitem{XIA2022204}
Xia, H., Zhao, B., Guo, P.: Synergetic learning structure-based neuro-optimal
  fault tolerant control for unknown nonlinear systems. Neural Networks
  \textbf{155},  204--214 (2022).
  \doi{https://doi.org/10.1016/j.neunet.2022.08.010},
  \url{https://www.sciencedirect.com/science/article/pii/S0893608022003100}

\bibitem{xu2020deep}
Xu, W., Parvin, H., Izadparast, H.: Deep learning neural network for
  unconventional images classification. Neural Processing Letters
  \textbf{52}(1),  169--185 (2020)

\bibitem{9445740}
Yin, Q., Xu, B., Zhou, K., Guo, P.: Bayesian pseudoinverse learners: From
  uncertainty to deterministic learning. IEEE Transactions on Cybernetics
  \textbf{52}(11),  12205--12216 (2022). \doi{10.1109/TCYB.2021.3079906}

\bibitem{8578294}
Zhang, J., Ghanem, B.: Ista-net: Interpretable optimization-inspired deep
  network for image compressive sensing. In: 2018 IEEE/CVF Conference on
  Computer Vision and Pattern Recognition. pp. 1828--1837 (2018).
  \doi{10.1109/CVPR.2018.00196}

\bibitem{7102696}
Zhang, Z., Xu, Y., Yang, J., Li, X., Zhang, D.: A survey of sparse
  representation: Algorithms and applications. IEEE Access  \textbf{3},
  490--530 (2015). \doi{10.1109/ACCESS.2015.2430359}

\bibitem{6937189}
Zhou, H., Huang, G.B., Lin, Z., Wang, H., Soh, Y.C.: Stacked extreme learning
  machines. IEEE Transactions on Cybernetics  \textbf{45}(9),  2013--2025
  (2015). \doi{10.1109/TCYB.2014.2363492}

\end{thebibliography}





\end{document}